\documentclass[conference]{IEEEtran}
\IEEEoverridecommandlockouts
\usepackage{cite}
\usepackage{amsmath,amssymb,amsfonts}
\usepackage{algorithmic}
\usepackage{graphicx}
\usepackage{textcomp}
\usepackage{xcolor}

\usepackage{multirow}
\usepackage{booktabs}
\usepackage{marvosym}

\def\BibTeX{{\rm B\kern-.05em{\sc i\kern-.025em b}\kern-.08em
    T\kern-.1667em\lower.7ex\hbox{E}\kern-.125emX}}
\begin{document}

\setlength{\textfloatsep}{5pt plus 1.0pt minus 2.0pt} 
\setlength{\floatsep}{5pt plus 1.0pt minus 2.0pt}     
\setlength{\intextsep}{5pt plus 1.0pt minus 2.0pt}    
\setlength{\abovedisplayskip}{3pt plus 1pt minus 1pt} 
\setlength{\belowdisplayskip}{3pt plus 1pt minus 1pt} 

\title{Disentangle-then-Refine: LLM-Guided Decoupling and Structure-Aware Refinement for Graph Contrastive Learning}

\author{
	Zhaoxing Li$^{1}$, Hai-Feng Zhang$^{2}$, Xiaoming Zhang$^{1,3,}$\textsuperscript{\Letter} \\
        
	$^{1}$Institute of Physical Science and Information Technology, Anhui University \\
        $^{2}$School of Mathematical Sciences, Anhui University \\
        $^{3}$Qinghai Institute of Science and Technology Information, Xining, China\\
        
        lzx@ahu.edu.cn,haifengzhang1978@gmail.com,\textsuperscript{\Letter}xmzhang@ustc.edu
}

\maketitle
\renewcommand{\thefootnote}{}
\footnotetext{\textsuperscript{\Letter} Corresponding Author. 

\thanks{This work was supported by the ``Kunlun Talent" Program of Qinghai.}
}


\begin{abstract}
Conventional Graph Contrastive Learning (GCL) on Text-Attributed Graphs (TAGs) relies on blind stochastic augmentations, inadvertently entangling task-relevant signals with noise. We propose \textbf{SDM-SCR}, a robust framework anchored in \textbf{Approximate Orthogonal Decomposition}. First, the \textbf{Semantic Decoupling Module (SDM)} leverages the instruction-following capability of Large Language Models (LLMs) to actively parse raw attributes into asymmetric, task-oriented signal and noise views. This shifts the paradigm from random perturbation to \textbf{semantic-aware disentanglement}. Subsequently, \textbf{Semantic Consistency Regularization (SCR)} exploits the spectral observation that semantic signals are topologically smooth while residual noise is high-frequency. SCR functions as a \textbf{selective spectral filter}, enforcing consistency only on the signal subspace to eliminate LLM hallucinations without over-smoothing. This ``Disentangle-then-Refine'' mechanism ensures rigorous signal purification. Extensive experiments demonstrate that SDM-SCR achieves SOTA performance in accuracy and efficiency.
\end{abstract}


\section{Introduction}
\label{intro}

Graph data have long been valued for their unique structural information, particularly Text-Attributed Graphs (TAGs), which encompass both complex topology and rich semantic information. While Graph Neural Networks (GNNs) have become the primary tool for graph learning \cite{tan2019deep}, the high cost of labeled data has shifted attention toward Graph Contrastive Learning (GCL) empowered by Self-Supervised Learning \cite{fang2024gaugllm}.

The core challenge of GCL lies in constructing high-quality views that preserve \textit{semantic invariance} while perturbing \textit{noise} \cite{Wu_Pan_Chen_Long_Zhang_Yu_2021}. In Computer Vision, this is effectively achieved through transformations like rotation or color desaturation \cite{mumuni2024survey}, which inherently maintain the identity of the object (e.g., a rotated plane remains a plane). However, graph data, being non-Euclidean, cannot directly adopt these methods \cite{liu2022graph}. Consequently, mainstream GCL approaches resort to stochastic augmentations such as random edge dropping or feature masking. Research \cite{li2023homogcl} indicates that these ``blind'' stochastic operations are theoretically flawed for TAGs because the definition of ``Signal'' vs. ``Noise'' is strictly task-dependent.

For instance, consider a product review graph where nodes are reviews. If the downstream task is \textit{Sentiment Analysis}, emotional adjectives (e.g., ``amazing", ``terrible") constitute the Signal. However, if the task is \textit{Product Categorization}, those same adjectives become Noise, and technical specifications (e.g., ``lens aperture", ``battery life") become the Signal. Random perturbation is agnostic to this distinction; it might inadvertently delete a key technical term crucial for categorization while preserving irrelevant sentiment words. This semantic misalignment prevents the model from learning optimal representations \cite{xia2022simgrace}.

\begin{figure}[htbp]
\centering
\includegraphics[width=1\linewidth]{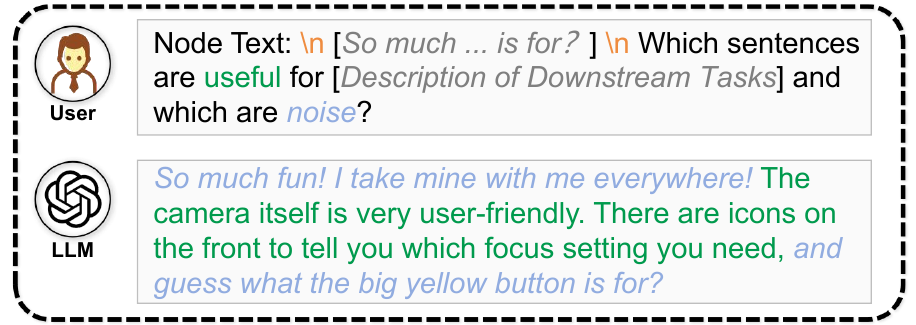}
\caption{ \label{Fig1} The text [\textit{So much fun...}] is from the Ele-Photo dataset. Each node’s text represents a comment for a product, and the model needs to perform a node classification task based on these comments. We use the \textcolor[rgb]{0.0,0.6039,0.3020}{\textbf{green font}} to indicate sentences related to the downstream task and the \textcolor[rgb]{0.6054,0.674,0.875}{\textbf{blue font}} to indicate noise sentences.}
\end{figure}

To address this, we propose that robust graph representation learning should strive for \textbf{Task-Aware Approximate Orthogonal Decomposition}. The core intuition is to encourage the decoupling of the embedding space into a \textit{semantic-dominant subspace} (Signal) and a \textit{noise-dominant subspace} (Noise), aiming for minimal mutual information between them. Since the definition of ``Signal" varies by task, static augmentations fail. Motivated by the impressive instruction-following and semantic reasoning capabilities of Large Language Models (LLMs), we propose the \textbf{Semantic Decoupling Module (SDM)}. Unlike random perturbations, SDM leverages LLMs as a dynamic semantic parser. As illustrated in Fig.~\ref{Fig1}, given a specific prompt for product classification, SDM explicitly separates the text into a \textit{Relevant View} (containing ``camera", ``user-friendly") and an \textit{Irrelevant View} (containing ``So much fun"). This allows the model to maximize agreement with the signal while explicitly pushing away the noise, reshaping the contrastive paradigm from random perturbation to semantic decoupling.

However, relying solely on instance-level LLM decoupling is insufficient. While our geometric verification confirms that SDM effectively cleaves the embedding space—evidenced by a sharp reduction in cosine similarity compared to random perturbations (see Appendix Fig 3), the observed non-zero residual similarity indicates that LLMs are not perfect oracles; they inevitably leak trace noise (hallucinations) into the relevant view, rendering the decoupling \textit{approximate} rather than strictly orthogonal. To suppress these residues, we introduce Semantic Consistency Regularization (SCR). Grounded in Graph Signal Processing, SCR leverages the observation that semantic signals are topologically smooth low-frequency whereas random noise is structurally independent high-frequency \cite{li2025ergnn}. By functioning as a selective spectral low-pass filter specifically on the decoupled Relevant View, SCR refines instance-level semantics without over-smoothing, thereby establishing a robust ``Disentangle-then-Refine'' workflow.

In summary, our contributions are as follows:
\begin{itemize}
    \item We propose the SDM, which leverages LLMs to construct asymmetric, task-oriented views. By instructing LLMs to parse content based on downstream objectives, SDM fundamentally shifts contrastive learning from reliance on blind random perturbations to genuine, task-specific semantic signals.
    \item We design SCR, a structure-aware filter grounded in the spectral properties of graph homophily. SCR is engineered to suppress residual noise inherited from LLMs by exploiting the observation that semantic signals are topologically smooth while noise is random. This prevents the over-smoothing of noise dimensions common in traditional GNNs.
    \item Extensive experiments across multiple datasets demonstrate that our method achieves SOTA performance while exhibiting significantly higher computational efficiency than competing approaches. Furthermore, we provide rigorous mathematical proofs to validate the theoretical soundness of our orthogonal decomposition framework.
\end{itemize}

\section{Related Work}

\subsection{Semantic Decoupling and View Construction}
\label{sec:related_sdm}

\subsubsection{From Random Perturbation to LLM Guidance} 
Early GCL frameworks, such as GRACE \cite{zhu2020deep} and MVGRL \cite{hassani2020contrastive}, heavily rely on stochastic input augmentations. They generate views via random edge dropping, node feature masking, or graph diffusion techniques like PageRank \cite{page1999pagerank}. While effective benchmarks, these methods operate under the assumption that noise is uniformly distributed, which is often flawed for TAGs. Randomly removing a keyword or an edge can inadvertently destroy semantic integrity or, conversely, leave structural noise intact \cite{ju2024towards}. To address the limitations of random augmentation, recent approaches have integrated LLMs to enhance feature quality. GIANT \cite{chien2022node} leverages XMC (Extreme Multi-label Text Classification) as a supervisory signal to refine node features. GAugLLM \cite{fang2024gaugllm} further utilizes LLMs to augment node sequences, achieving robust performance. Similarly, LATEX-GCL \cite{yang2024latex} instructs LLMs to perform summarization and rewriting tasks to generate augmented views.

\subsubsection{The Need for Orthogonal Decomposition}
Although these LLM-based methods outperform stochastic baselines, they face two critical issues: 
(1) \textit{Computational Inefficiency}: Methods like GAugLLM and LATEX-GCL require multiple LLM inference passes per node, incurring prohibitive costs for large-scale graphs.
(2) \textit{Entangled Representations}: They typically treat LLM outputs as enriched features without explicitly separating semantic signals from noise. Recent studies on disentangled graph representation \cite{xie2024graph, sasaki2025benchmarking} suggest that effective learning requires identifying and decomposing underlying factors. Inspired by this, our \textbf{SDM} departs from simple augmentation. Instead, it employs LLMs to perform an efficient, instance-level Approximate Orthogonal Decomposition (AOD), projecting raw text into asymmetric \textit{Relevant} (Signal) and \textit{Irrelevant} (Noise) subspaces.

\subsection{Spectral Filtering and Structural Regularization}
\label{sec:related_scr}

\subsubsection{Contrastive Mechanisms and Homophily}
Standard GCL methods optimize pairwise agreements using objectives like InfoNCE, an implicit semantic-alignment objective \cite{wang2025infonce}. BGRL \cite{thakoor2021large} introduces an asymmetric online-target encoder architecture updated via Exponential Moving Average (EMA)  to avoid collapse without negative sampling. A key trend in recent years is incorporating structural priors into the contrastive objective. HomoGCL \cite{li2023homogcl} explicitly utilizes the homophily assumption, treating neighbors as positive samples to guide representation learning. Similarly, LanSAGNN \cite{li2025refine} leverages language semantics to refine structural interactions, mitigating interference from irrelevant neighbors.

\subsubsection{Spectral Graph Filtering}
However, relying solely on spatial neighbors can lead to the ``over-smoothing" problem, where node representations become indistinguishable in deep layers or noisy graphs. Recent theoretical advances in Spectral Graph Neural Networks (SGNNs) emphasize the importance of tailored graph filters. For instance, Li et al. \cite{li2025ergnn} proposed ERGNN with explicitly optimized rational filters to better handle high-frequency noise. Furthermore, entropy-based analyses \cite{qian2025exploring} have shown that preserving high-entropy features is crucial for mitigating over-smoothing in graph classification. Existing methods like PolyGCL \cite{chen2024polygcl} attempt to use frequency filters but often apply them indiscriminately to the entire feature set. This leads to the ``blind smoothing" of noise. Our \textbf{SCR} module bridges the gap between disentanglement and spectral filtering. By leveraging the decoupled views from SDM, SCR applies a structure-aware low-pass filter specifically to the \textit{Relevant} subspace, while allowing the \textit{Irrelevant} subspace to absorb high-frequency spectral residue. This design aligns with the latest findings in robust graph filtering \cite{li2025ergnn, liu2024fine}, ensuring that topological smoothing reinforces semantics without propagating noise.

\begin{figure*}[htbp]
\centering
\resizebox{1.3\columnwidth}{!}{
\includegraphics[width=1\linewidth]{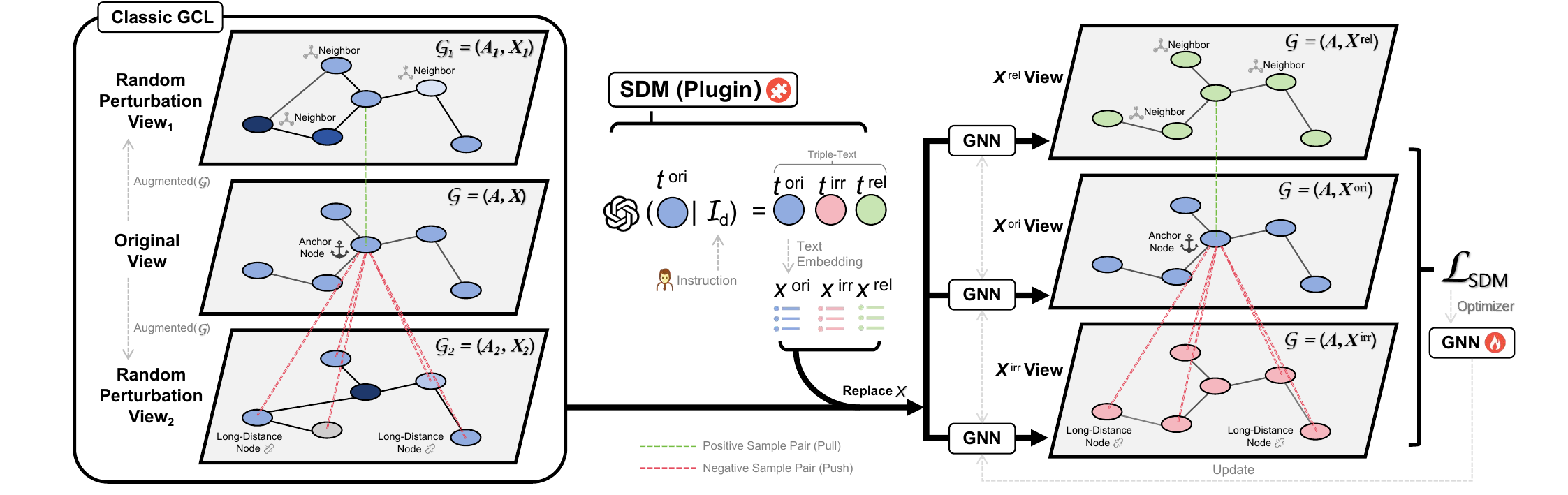}
}
\caption{ \label{Fig2} The Difference Between Traditional Methods' Data Augmentation and SDM Module.}
\end{figure*}

Our \textbf{SCR} module fundamentally reconceptualizes topological regularization as a \textbf{precision spectral gating mechanism}. Rather than applying a monolithic filter, SCR synergizes with SDM to treat the graph structure as a \textbf{selective semantic amplifier}: it enforces rigorous low-pass consistency exclusively within the \textit{Relevant} subspace to distill shared semantics, while strategically utilizing the \textit{Irrelevant} subspace as a ``spectral sink'' to sequester high-frequency noise. This design breaks the trade-off between smoothness and distinctiveness found in traditional GNNs \cite{li2025ergnn, liu2024fine}, establishing a robust \textbf{Disentangle-then-Refine} paradigm where structural propagation is strictly governed by semantic validity.

\section{Methodology}
\label{pipel}

\subsection{Formulation and Approximate Orthogonal Decomposition}
Given a TAG $\mathcal{G} = (T, A)$ with $N$ nodes, where $T = \{t^{ori}_i\}_{i=1}^N$ denotes the set of raw text attributes and $A \in \{0,1\}^{N \times N}$ is the adjacency matrix. The goal is to learn a GNN-based encoder $\mathcal{E}(\cdot)$ that obtains the final representation for node $i$ based on the embedding $\mathbf{x}_i$ derived from its raw text $t_i$.

We posit that the information within the raw embedding $\mathbf{x}_i^{ori}$ can be ideally modeled via an \textbf{AOD} principle:
\begin{equation}
    \mathbf{x}_i^{ori} = \mathbf{s}_i + \mathbf{n}_i, \quad \text{aiming for} \quad \langle \mathbf{s}_i, \mathbf{n}_i \rangle \to 0
\end{equation}
where $\mathbf{s}_i$ represents the \textbf{Semantic Signal} (features causally related to the label) and $\mathbf{n}_i$ represents the \textbf{Noise} (task-irrelevant patterns). Traditional GCL methods inadvertently entangle these components because random perturbations treat signal and noise symmetrically, and thus often destroy label-causal semantics together with nuisance factors. While strict orthogonality is \textbf{impossible} in complex semantic spaces, this formulation serves as the theoretical guiding objective for our approximation strategy. In contrast, our framework explicitly exploits LLMs to obtain \emph{semantically guided} views that approximate this decomposition, which is intrinsically more informed than random augmentation.

\subsection{Semantic Decoupling Module (SDM)}
The SDM aims to operationalize the AOD principle by leveraging the semantic reasoning capabilities of LLMs as a ``soft'' decomposition operator. Rather than relying on blind perturbations, SDM uses LLMs to produce views that are \emph{signal-dominant} and \emph{noise-dominant}, respectively, yielding a substantially cleaner separation than random augmentation despite remaining imperfections. 

As illustrated in Fig.~\ref{Fig2}, SDM functions as a versatile \textbf{plug-in} for existing GCL frameworks. Instead of modifying the GNN encoder structure, it simply intercepts the input stage: the raw node attributes $t^{\text{ori}}$ are processed by the LLM to generate disentangled views, which then \textit{replace} the original noisy features (as denoted by the ``Replace $X$" path in the figure). This design allows the SDM to upgrade many existing GCL frameworks on TAGs.

\subsubsection{Instruction-Driven Decomposition with Residual Awareness}
Instead of random augmentation, we input a task-specific instruction $\mathcal{I}_\text{d}$ and the raw text $t^{\text{ori}}_i$ into an LLM. The instruction directs the LLM to parse the content, yielding two distinct textual components:
\begin{equation}
    LLM(t_i^{\text{ori}} \mid \mathcal{I}_\text{d}) \rightarrow (t_{i}^{\text{rel}}, t_{i}^{\text{irr}}).
\end{equation}
Here, $t_{i}^{\text{rel}}$ denotes the task-relevant content, while $t_{i}^{\text{irr}}$ captures the task-irrelevant noise. We pass the resulting text set $\{t_i^{\text{ori}}, t_i^{\text{rel}}, t_i^{\text{irr}}\}$ through a text embedding encoder to obtain their corresponding embeddings. These are formally defined with residual terms as follows:
\begin{align}
    \mathbf{x}_i^{ori} &= \mathbf{s}_i + \mathbf{n}_i, \\
    \mathbf{x}_i^{rel} &= \mathbf{s}_i + \epsilon_i \quad \text{(signal-dominant view)}, \\
    \mathbf{x}_i^{irr} &= \mathbf{n}_i + \zeta_i \quad \text{(noise-dominant view)}.
\end{align}
Here, $\epsilon_i$ represents residual noise or hallucinations introduced by the LLM, and $\zeta_i$ denotes trace semantics leaked into the irrelevant view. Although modern LLMs are not perfect semantic oracles, their strong language understanding is sufficient in our setting to provide a coarse semantic split, where $\mathbf{x}_i^{rel}$ tends to align with task-relevant signal while $\mathbf{x}_i^{irr}$ predominantly captures nuisance factors. Crucially, $\mathcal{I}_\text{d}$ incorporates only task background to resolve signal ambiguity, ensuring that no instance-level ground truth labels are exposed during this phase.

\subsubsection{Asymmetric Contrastive Optimization}
To approximate this orthogonal separation in the embedding space, SDM employs an asymmetric contrastive objective. Although $\mathbf{x}_k^{\text{irr}}$ may still contain trace semantic residuals (as shown in ablation studies), it is \textbf{dominated by the noise subspace}. Pushing away $\mathbf{x}_k^{\text{irr}}$ enforces a stricter constraint, compelling the encoder to filter out nuisance factors and focus exclusively on the discriminative features in $\mathbf{x}_i^{\text{rel}}$. Crucially, since SDM is designed to function as a versatile plug-in compatible with various GCL backbones, we exclusively utilize the irrelevant view as the negative sample to avoid interference with the instance discrimination mechanisms inherent in the host frameworks.

For any anchor node $i$, the SDM loss is defined as:
\begin{equation}
    \mathcal{L}_i^{\text{SDM}} = -\log \left[ \frac{\exp \left( \frac{\text{sim}(\mathbf{x}_i^{\text{ori}}, \mathbf{x}_i^{\text{rel}})}{\tau} \right)}{\exp \left( \frac{\text{sim}(\mathbf{x}_i^{\text{ori}}, \mathbf{x}_i^{\text{rel}})}{\tau} \right) + \sum_{k \neq i} \exp \left( \frac{\text{sim}(\mathbf{x}_i^{\text{ori}}, \mathbf{x}_k^{\text{irr}})}{\tau} \right)} \right],
\end{equation}
where $\text{sim}(\cdot, \cdot)$ denotes cosine similarity and $\tau$ is the temperature. The final objective $\mathcal{L}^{\text{SDM}}$ is obtained by averaging $\mathcal{L}_i^{\text{SDM}}$ over all $N$ nodes. This mechanism acts as a ``semantic magnet,'' pulling $\mathbf{x}_i^{\text{ori}}$ towards the signal subspace $\mathbf{s}_i$ (approximated by $\mathbf{x}_i^{\text{rel}}$) while expelling the noise components encoded in $\mathbf{x}_k^{\text{irr}}$.

\subsection{Semantic Consistency Regularization (SCR)}
Although SDM achieves instance-level decoupling, LLMs are inherently stochastic and not perfect oracles. As defined in Eq.~(4), the relevant view $\mathbf{x}_i^{\text{rel}} = \mathbf{s}_i + \epsilon_i$ still contains a residual noise term $\epsilon_i$. If left unaddressed, contrastive learning may overfit to these hallucinations. To further suppress $\epsilon_i$ in a \emph{structure-aware} manner, we introduce SCR to exploit the spectral properties of the graph topology.

\subsubsection{Mitigating Residuals via Spectral Filtering}
The core insight of SCR is based on the distinct spectral characteristics of the signal $\mathbf{s}$ and the residual $\epsilon$:
\begin{itemize}
    \item \textbf{Signal ($\mathbf{s}$):} Based on the \textbf{structural consistency prior}, valid semantic signals tend to exhibit \textbf{low-frequency} characteristics across the local topology (i.e., $\mathbf{s}_i \approx \mathbf{s}_j$ for connected nodes in reliable neighborhoods).
    \item \textbf{Residual ($\epsilon$):} Generated largely independently by the LLM, these errors manifest as high-frequency noise. Recent findings \cite{nonkes2024leveraging} corroborate that hallucinations are structurally inconsistent (i.e., random relative to topology); thus, our SCR can effectively suppress them through low-pass filtering.
\end{itemize}

SCR acts as a \textbf{variance reduction operator}. By enforcing consistency on $\mathbf{x}^{\text{rel}}$ within the neighborhood $\mathcal{N}(i)$, we implicitly perform a Laplacian-like smoothing operation. This mechanism is architecture-agnostic and applies to any Message Passing Neural Network (MPNN) framework. For a local neighborhood, the aggregated representation approximates:
\begin{equation}
    \frac{1}{|\mathcal{N}(i)|} \sum_{j \in \mathcal{N}(i)} \mathbf{x}_j^{\text{rel}} = \underbrace{\frac{1}{|\mathcal{N}(i)|} \sum_{j} \mathbf{s}_j}_{\approx \mathbf{s}_i \text{ (signal)}} + \underbrace{\frac{1}{|\mathcal{N}(i)|} \sum_{j} \epsilon_j}_{\text{variance-reduced residual}}.
\end{equation}
Under mild independence and zero-mean assumptions on $\epsilon_j$, the second term exhibits reduced variance as the neighborhood size grows, consistent with the classical law of large numbers. While illustrated here using mean aggregation (typical of GraphSAGE or GCN), this variance reduction effect generalizes to other aggregation schemes, where the weighted sum of independent noise terms similarly tends towards zero~\cite{cong2020minimal,vatter2024size}. Thus, SCR effectively filters out LLM-induced residuals without requiring ground-truth supervision, while reinforcing structurally consistent semantic signals.

Remarkably, this design also confers robustness against structural noise. Unlike traditional GNNs that blindly aggregate all neighbor features, SCR performs smoothing exclusively on the purified $x^{rel}$, effectively blocking the propagation of task-irrelevant noise across potentially unreliable connections. We provide a theoretical proof in the Appendix demonstrating that this mechanism yields a lower error bound than standard message passing, even under heterophilic settings.

\subsubsection{The Irrelevant View as a Spectral Sink}
Regarding the irrelevant view $\mathbf{x}^{\text{irr}} = \mathbf{n}_i + \zeta_i$, we deliberately \textbf{exclude} it from topological regularization. Since the leakage term $\zeta_i$ (trace semantics) and structural noise $\mathbf{n}_i$ are typically high-frequency or heterophilic, enforcing smoothness on $\mathbf{x}^{\text{irr}}$ would result in \emph{negative transfer}, propagating noise to neighbors. By leaving $\mathbf{x}^{\text{irr}}$ unconstrained, we treat the noise subspace as a \textbf{spectral sink}, allowing the model to offload non-smooth, high-frequency variations (including $\epsilon$ and $\mathbf{n}$) into this orthogonal subspace, thereby purifying the relevant view.

\subsubsection{Regularization Objective}
Formally, SCR minimizes the cosine distance of the relevant views between connected nodes, acting as a structure-aware low-pass filter:
\begin{equation}
    \mathcal{L}^{\text{SCR}} = \frac{1}{N} \sum_{i=1}^{N} \frac{1}{|\mathcal{N}(i)|} \sum_{j \in \mathcal{N}(i)} \left( 1 - \text{sim}(\mathbf{x}_i^{\text{rel}}, \mathbf{x}_j^{\text{rel}}) \right).
\end{equation}
Both $\mathcal{L}^{\text{SDM}}$ and $\mathcal{L}^{\text{SCR}}$ are node-averaged scalar objectives. While their absolute magnitudes may differ due to the logarithmic nature of the contrastive loss versus the linear nature of the cosine regularization, they are balanced via the trade-off hyperparameter $\lambda$, ensuring consistent gradient optimization.

\subsection{Overall Objective}
To strictly balance the contribution of instance-level decoupling (which relies on LLM priors) and structure-level refinement (which relies on topological priors), we formulate the final objective as a convex combination:
\begin{equation}
    \mathcal{L} = \lambda \mathcal{L}^{\text{SDM}} + (1 - \lambda) \mathcal{L}^{\text{SCR}},
\end{equation}
where $\lambda \in [0,1]$ is a trade-off hyperparameter. Since both terms are dimensionless, node-averaged contrastive or regularization losses, $\lambda$ directly controls the relative emphasis between semantic decoupling and spectral refinement, allowing the model to leverage the semantic richness of LLMs while using the graph structure to correct their inevitable hallucinations.

\begin{table*}
\centering
\resizebox{1.5\columnwidth}{!}{
\renewcommand\arraystretch{0.3}
\begin{tabular}{c|c|c|ccc|cc}
\toprule
\textbf{Category} & \textbf{Method} & \textbf{Input} & \textbf{Citeseer}& \textbf{Pubmed}& \textbf{Wiki-CS}& \textbf{Ele-Photo$^{*}$} & \textbf{Books-History$^{*}$} \\ \midrule
\multirow{8}{*}{Classic and SOTA GCL} & Grace         & $X,A$   & 70.80±0.52 & 80.21±0.29 & 78.27±0.14 & -          & -          \\
                                  & DGI           & $X,A$   & 71.82±0.70 & 76.80±0.60 & 75.35±0.14 & -          & -          \\
                                  & MVGRL         & $X,A$   & 73.02±0.43 & 80.12±0.51 & 77.56±0.10 & -          & -          \\
                                  & ProGCL        & $X,A$   & 69.78±0.48 & 79.20±0.33 & 78.68±0.12 & -          & -          \\
                                  & HomoGCL       & $X,A$   & 72.18±0.62 & 81.04±0.37 & 79.12±0.07 & -          & -          \\
                                  & PolyGCL       & $X,A$   & 72.54±0.59 & 80.51±0.76 & 75.54±0.37 & -          & -          \\
                                  & IFL-GC        & $X,A$   & 71.72±0.44 & 80.93±0.21 & 79.25±0.28 & -          & -          \\ \midrule
                                  & GIANT         & $T,A$   & 72.96±0.41 & 81.54±0.70 & 79.34±0.25 & 74.85±0.41 & 74.41±0.73 \\ 
\multirow{4}{*}{LM-Based}         & LLM-ZeroShot  & $T$     & 49.59      & \textbf{85.47}      & 66.81      & 43.58      & 51.04      \\
                                  & LLM-Embedding & $T,A$   & 70.81±1.45 & 79.31±1.19 & 79.51±0.46 & 74.68±0.23 & 77.03±0.33 \\
                                  & LATEX-GCL     & $T,A$   & 71.39±0.74 & 80.75±1.28 & 80.29±0.68 & 76.81±0.24 & 78.26±0.27 \\
                                  & GAugLLM       & $T,A$   & {\textbf{73.71±0.43}} & 82.90±0.62 & {\underline{82.11±0.47}} & 77.69±0.21 & 77.98±0.32 \\ \midrule
\multirow{2}{*}{Ours}             & SDM           & $T,A$   & 73.54±0.64 & 82.97±0.84 & 81.93±0.21 & {\underline{78.25±0.14}} & {\underline{78.84±0.18}} \\
                                  & \textbf{SDM - SCR}     & $T,A$   & \underline{73.70±0.93} & {\underline{83.03±0.52}} & \textbf{82.18±0.60} & \textbf{78.89±0.21} & \textbf{78.93±0.15} \\ \bottomrule
\end{tabular}
    }
\caption{Comparison of SDM-SCR framework with other methods. Ele-Photo$^{*}$ and Books-History$^{*}$, two TAGs datasets, are not included in common GNNs libraries such as PyG and DGL, and thus lack official standard embeddings. }
\label{table1}
\end{table*}

\begin{table}[htbp]
  \centering
  \small
  \renewcommand\arraystretch{0.2}
  \resizebox{0.75\linewidth}{!}{
  \begin{tabular}{@{}c |l| c| c@{}}
    \toprule
    \textbf{Category} & \textbf{Component} & \textbf{Citeseer} & \textbf{Wiki-CS} \\
    \midrule
    \multirow{6}{*}{SDM Plug-in}
        & GBT & 71.09$\pm$0.45 & 80.21$\pm$0.38 \\
        & GBT (+ SDM) & 71.43$\pm$0.88 & 80.27$\pm$0.42 \\
        & Grace & 71.11$\pm$0.76 & 80.55$\pm$0.16 \\
        & Grace (+ SDM) & 71.56$\pm$1.10 & 81.06$\pm$0.48 \\
        & HomoGCL & 73.17$\pm$0.34 & 81.56$\pm$0.58 \\
        & HomoGCL (+ SDM) & 73.51$\pm$0.53 & 81.91$\pm$0.19 \\
    \midrule
    \multirow{3}{*}{Subspace}
        & Embedding ($X^{ori}$) & 70.81$\pm$1.42 & 79.51$\pm$0.46 \\
        & Embedding ($X^{rel}$) & 71.41$\pm$0.61 & 80.00$\pm$0.32 \\
        & Embedding ($X^{irr}$) & 65.67$\pm$1.11 & 77.37$\pm$0.31 \\
    \bottomrule
  \end{tabular}
  }
  \caption{Ablation Study of SDM. We test SDM as a plug-in for different GCL frameworks and evaluate the quality of the decomposed subspaces.}
  \label{table2}
\end{table}

\begin{table}[htbp]
  \centering
  \small
  \renewcommand\arraystretch{0.2}
  \resizebox{0.75\linewidth}{!}{
  \begin{tabular}{l | c | c}
    \toprule
    \textbf{LLM Backbone} & \textbf{Citeseer} & \textbf{Wiki-CS} \\ \midrule
    Llama-3.1-8B & 73.70 $\pm$ 0.93 & 82.18 $\pm$ 0.60 \\
    DeepSeek-7B & 73.10 $\pm$ 0.33 & 82.38 $\pm$ 0.41 \\
    Mistral-0.3-7B & 73.63 $\pm$ 0.48 & 82.06 $\pm$ 0.21 \\
    GPT-4o-mini & 73.86 $\pm$ 0.61 & 82.75 $\pm$ 0.29 \\
    Qwen-3-4B & 73.20 $\pm$ 0.64 & 81.50 $\pm$ 0.41 \\
    Gemma-3-1B & 72.40 $\pm$ 0.32 & 80.84 $\pm$ 0.26 \\
    \bottomrule
  \end{tabular}
  }
  \caption{Performance robustness of SDM-SCR across different LLMs.}
  \label{table3}
\end{table}
·

\section{Experiment}
\label{exper}

We conduct extensive testing of the proposed \textbf{SDM} and \textbf{SCR} modules across datasets from multiple domains, including Citeseer \cite{sen2008collective}, Wiki-CS \cite{mernyei2020wiki}, Pubmed \cite{Sen_Namata_Bilgic_Getoor_Galligher_Eliassi}, Ele-Photo, and Books-History \cite{Ni_Li_McAuley_2019}. For details on the dataset information, experimental setup, code, and other related content, please refer to the supplementary materials. To facilitate comparisons with existing GCL schemes, we use mainstream contrastive learning dataset splits for the node classification problem. Specifically, this section addresses the following research questions:
\textbf{RQ1}: How does the SDM-SCR framework perform compared to state-of-the-art GCL methods? 
\textbf{RQ2}: Can the SDM effectively decouple semantic signals from noise? How does it contribute to performance when used as a plug-in for other frameworks?
\textbf{RQ3}: Does the proposed framework demonstrate robust performance across diverse LLM backbones with varying parameter scales and architectures?

\subsection{Main Experiment (RQ1)}
We compare our proposed modules against advanced methods using the common GCL evaluation scheme: labels are unavailable during the representation learning phase and are only used post-training for node classification tasks via a logistic regression classifier. To ensure fairness, all LLM-based methods are replicated using the same LLM (Llama3.1-8b \cite{dubey2024llama}). The results are reported in Tab.~\ref{table1}. SDM-SCR represents our full ``Disentangle-then-Refine" framework. 

As shown in Tab.~\ref{table1}, the SDM module alone already achieves competitive performance against existing LLM-based methods (e.g., GAugLLM), verifying the effectiveness of the AOD strategy. Furthermore, the complete SDM-SCR framework achieves SOTA results, demonstrating that the spectral filtering mechanism of SCR successfully purifies the semantic subspace by removing high-frequency residual noise.

\subsection{Ablation Study: Effectiveness of SDM (RQ2)}

In this section, we investigate the contribution of the SDM. SDM is designed to replace stochastic perturbations with semantic-aware subspace decomposition. To verify its versatility and the validity of the decomposition hypothesis, we conduct two sets of experiments presented in Tab.~\ref{table2}.

\subsubsection{SDM as a General Plug-in} 
The `SDM Plug-in' section in Tab.~\ref{table2} demonstrates using SDM as a plug-in to enhance other GCL frameworks. By replacing the random augmentation (e.g., edge dropping) in these frameworks with the Relevant/Irrelevant views generated by SDM, we observe consistent performance gains. This proves that the semantic signal $\mathbf{s}$ extracted by SDM is universally more robust than the noisy features preserved by random perturbations.

\subsubsection{Subspace Quality Analysis}
To further verify the quality of the decoupled subspaces and the necessity of our contrastive paradigm, we directly evaluate the classification capability of embeddings \emph{without} any GCL training, comparing the original features $X^{ori}$ with the relevant subspace $X^{rel}$ and the irrelevant/noise subspace $X^{irr}$. The results show that $X^{rel}$ consistently matches or surpasses $X^{ori}$, whereas $X^{irr}$ performs significantly worse, confirming that SDM concentrates discriminative semantics into the relevant view. Meanwhile, simply using $X^{rel}$ as an ``LLM-Embedding'' plug-in baseline still lags behind the full SDM-SCR framework, indicating that the asymmetric contrastive repulsion is crucial to further align semantic signals and explicitly separate noise via the safe negative $X^{irr}$. Overall, these findings demonstrate both the effectiveness of our decoupling and the advantage of SDM-SCR as a lightweight and effective plug-in for existing GCL pipelines.

\subsection{LLM Robustness Analysis (RQ3)}
We evaluate the robustness of our framework by instantiating the SDM module with six different LLM backbones. As demonstrated in Tab.~\ref{table3}, SDM-SCR exhibits remarkable \textbf{universal effectiveness}: even when employing the lightweight Gemma-3-1B, our method maintains high performance. This indicates that the core decoupling mechanism operates independently of the model scale. Furthermore, open-source models such as Llama-3.1-8B achieve results competitive with closed-source GPT models, confirming that SDM-SCR serves as a cost-effective and model-agnostic plug-in.

\section{Conclusion}
In this paper, we propose SDM-SCR, a novel graph contrastive learning framework that leverages the semantic understanding capabilities of LLMs to achieve task-oriented decoupling of signal and noise, and further utilizes the spectral properties of graph homophily for structure-aware refinement. Extensive experiments confirm that SDM-SCR achieves SOTA performance with significantly reduced inference time compared to existing LLM-enhanced methods. Furthermore, our framework demonstrates remarkable robustness across diverse mainstream LLM backbones and embedding initialization methods. The performance of SDM applied as a plug-in across different contrastive learning frameworks proves its potential as a universal enhancer for graph representation learning. To ensure reproducibility and theoretical rigor, we provide comprehensive supplementary materials in the Appendix, covering runtime comparisons, additional comparative experiments, dataset details, experimental environments, LLM instructions, and rigorous mathematical proofs validating the efficacy of our orthogonal decomposition and variance reduction mechanisms.

\bibliographystyle{IEEEbib}
\bibliography{bibicme2026references}


\end{document}